\documentclass[letterpaper, 10 pt, conference]{ieeeconf}  

\IEEEoverridecommandlockouts                              
\usepackage{amsfonts}
\usepackage{mathtools}
\usepackage{amsmath}
\usepackage{amssymb}
\usepackage{framed}
\usepackage{balance}
\usepackage{breqn}
\usepackage{subfigure}
\usepackage{graphics} 
\usepackage{lipsum}
\usepackage{graphicx}
\usepackage{mathptmx} 
\usepackage{fancyhdr}
\usepackage[nospace]{cite}
\usepackage[ruled,vlined]{algorithm2e}
\usepackage{setspace}
\usepackage{amssymb}  
\usepackage{algpseudocode}
\usepackage{booktabs}
\usepackage{tabularx}
\usepackage{booktabs}
\usepackage{array}
\usepackage{etex}
\usepackage{adjustbox}
\usepackage{adjustbox,lipsum}
\usepackage[table]{xcolor}
\usepackage{gensymb}
\usepackage{nicefrac}
\usepackage{mdwtab}
\usepackage{multirow}
\usepackage{colortbl}
\usepackage{hyperref}
\usepackage{color,soul}
\usepackage{epsfig} 
\usepackage{graphics} 

\DeclareMathAlphabet{\mathcal}{OMS}{cmsy}{m}{n}

\usepackage{stmaryrd}

\algrenewcommand\algorithmicforall{\textbf{foreach}}
\algrenewcommand\algorithmicindent{.8em}

\usepackage{accents}
\newcommand*{\dt}[1]{\accentset{\mbox{\bfseries .}}{#1}}
\newcommand*{\ddt}[1]{\accentset{\mbox{\bfseries .\hspace{-1.0ex}.}}{#1}}

\DeclareMathOperator*{\argmax}{arg\,max}  

\begin{document}

\title{\LARGE \bf
Gaussian-Process-based Robot Learning from Demonstration}
\author{Miguel Arduengo$^{1}$, Adri\`a Colom\'e$^{1}$, Joan Lobo-Prat$^{1}$, Luis Sentis$^{2}$ and Carme Torras$^{1}$
\thanks{This work is partially funded by ERC Advanced Grant H2020-741930 (project CLOTHILDE).}
\thanks{$^{1}$Institut de Robotica i Informatica Industrial (IRI), Barcelona, Spain. \{marduengo, acolome, jlobo, torras\}@iri.upc.edu}
\thanks{$^{2}$Human Centered Robotics Laboratory (HCRL), UT Austin, USA. lsentis@austin.utexas.edu}}

\maketitle
\thispagestyle{empty}
\pagestyle{empty}


\begin{abstract}

Endowed with higher levels of autonomy, robots are required to perform increasingly complex manipulation tasks. Learning from demonstration is arising as a promising paradigm for transferring skills to robots. It allows to implicitly learn task constraints from observing the motion executed by a human teacher, which can enable adaptive behavior. We present a novel Gaussian-Process-based learning from demonstration approach. This probabilistic representation allows to generalize over multiple demonstrations, and encode variability along the different phases of the task. In this paper, we address how Gaussian Processes can be used to effectively learn a policy from trajectories in task space. We also present a method to efficiently adapt the policy to fulfill new requirements, and to modulate the robot behavior as a function of task variability. This approach is illustrated through a real-world application using the TIAGo robot.

\end{abstract} 

\section{INTRODUCTION}

In the context of robotics, learning from demonstration (LfD) is the paradigm in which robots learn a task policy from examples provided by a human teacher. This facilitates non-expert robot programming, since task constraints and requirements are learned implicitly from the demonstrated motion, which can enable adaptive behavior \cite{Ravichandar2020}. 

\begin{figure}[t]
    \centering
    {\includegraphics[width=0.95\linewidth]{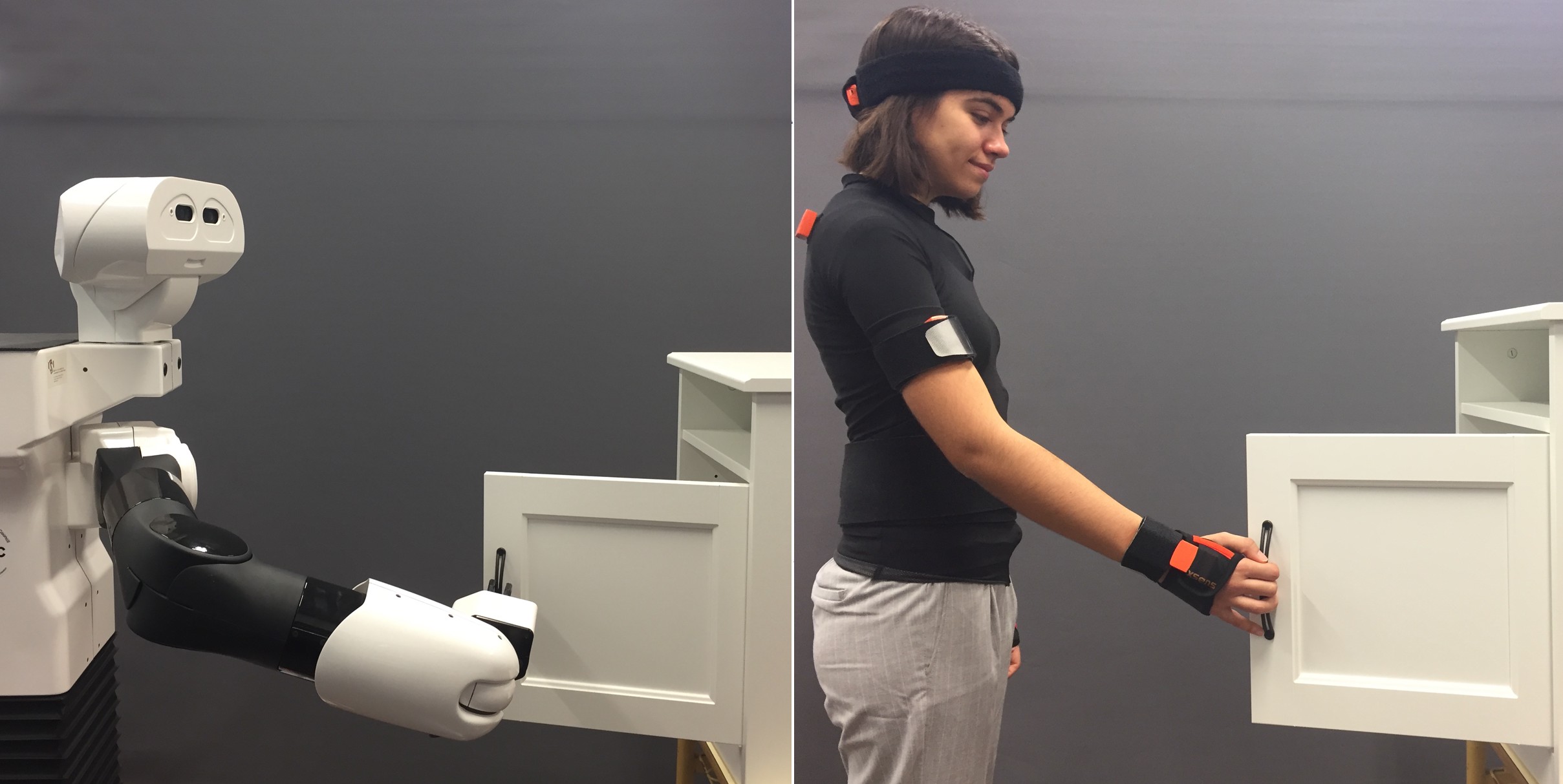}}
    \caption{The proposed Gaussian-Process-based learning from demonstration approach allows to teach the robot manipulation tasks such as opening doors.}
    \label{fig1}
    \vspace{-1.8em}
\end{figure}

Trajectory-based methods are a well-established approach to learn movement policies in robotics  \cite{Colome2020}. These methods encode skills by extracting trajectory patterns from demonstrations (Figure \ref{fig1}), using a variety of techniques to retrieve a generalized shape of the trajectory. Over the past two decades, it has been an intensive field of study. Among the most relevant contributions, the following methods can be highlighted: Dynamic Movement Primitives (DMP) \cite{Ijspeert2001,Pastor2009}, Probabilistic Movement Primitives (ProMP) \cite{Paraschos2018}, Gaussian Mixture Model-Gaussian Mixture Regression (GMM-GMR) \cite{Calinon2016,Pignat2019}, Kernelized Movement Primitives \cite{Huang2019b,Huang2019c}, and Gaussian Processes (GP)\cite{Nguyen2008,Forte2010}. These representations have proved successful at learning and generalizing trajectories. However, each model presents its strengths and shortcomings.

The main advantage of probabilistic-based methods (GMM-GMR, ProMP, KMP and GP) is that they not only retrieve an estimate of the underlying trajectory across multiple demonstrations, but also encode its variability by means of a covariance matrix. This information, which can be inferred from the dispersion of the collected data, can be exploited for the execution of the task, i.e., specifying the robot tracking precision or switching the controller \cite{Silverio2018}.

\begin{table*}[t]
\footnotesize
\def\arraystretch{1.5}
\setlength\tabcolsep{1.8mm}
\centering
\caption{Comparison among the state-of-the-art and our approach}
\label{Table1}
\begin{tabular}{||c||c|c|c|c|c|c|c||}
    \cline{2-8}
    \multicolumn{1}{c||}{} & \textbf{DMP} \cite{Ijspeert2001,Koutras2019} & \textbf{ProMP} \cite{Paraschos2018} & \textbf{GMM-GMR} \cite{Calinon2016,Zeestraten2017} & \textbf{GP} \cite{Schneider2010a,Umlauft2017} & \textbf{KMP} \cite{Huang2019b,Huang2019c} &  \textbf{GMM-GP} \cite{Jaquier2019} & \textbf{Our approach} \\ [0.5ex] 
    \hline\hline
    \textit{High-dimensional Learning} & $\normalsize{-}$ & $\normalsize{-}$ & $\normalsize{\checkmark}$ & $\normalsize{\checkmark}$ & $\normalsize{\checkmark}$ & $\normalsize{\checkmark}$ & $\normalsize{\checkmark}$ \\
    \hline
    \textit{Via-point Adaptation} & $\normalsize{-}$ & $\normalsize{\checkmark}$ & $\normalsize{-}$ & $\normalsize{-}$ & $\normalsize{\checkmark}$ & $\normalsize{\checkmark}$ & $\normalsize{\checkmark}$ \\ 
    \hline
    \textit{Task Variability} & $\normalsize{-}$ & $\normalsize{\checkmark}$ & $\normalsize{\checkmark}$ & $\normalsize{\checkmark}$ & $\normalsize{\checkmark}$ & $\normalsize{\checkmark}$ & $\normalsize{\checkmark}$ \\ 
    \hline
    \textit{Prediction Uncertainty} & $\normalsize{-}$ & $\normalsize{-}$ & $\normalsize{-}$ & $\normalsize{\checkmark}$ & $\normalsize{\checkmark}$ & $\normalsize{\checkmark}$ & $\normalsize{\checkmark}$ \\ 
    \hline
    \textit{Prior Information} & $\normalsize{-}$ & $\normalsize{-}$ & $\normalsize{-}$ & $\normalsize{\checkmark}$ & $\normalsize{-}$ & $\normalsize{\checkmark}$ & $\normalsize{\checkmark}$ \\ 
    \hline
    \textit{Task Space Rotations} & $\normalsize{\checkmark}$ & $\normalsize{\checkmark}$ & $\normalsize{\checkmark}$ & $\normalsize{-}$ & $\normalsize{\checkmark}$ & $\normalsize{-}$ & $\normalsize{\checkmark}$ \\ 
    \hline
    
\end{tabular}
\vspace{-4mm}
\end{table*} \normalsize

Unlike probabilistic-based methods, at the cost of not encoding the variability of the task, DMP only requires a single demonstration. Generalization is achieved by assuming trajectories to be solutions of a deterministic dynamical system, achieving remarkable success. A drawback of DMP, and also ProMP, is that they rely on the manual specification of basis functions, which requires expert knowledge and makes the learning problem with high-dimensional inputs almost intractable. GMM-GMR, in contrast, has proven successful in handling this kind of demonstrations. KMP and GP, due to their kernel treatment, can be implemented for manipulation tasks where high-dimensional inputs and outputs are required.

In LfD, it is also interesting to transfer the learned motion to unseen scenarios while maintaining the general trajectory shape as in the demonstrations. By exploiting the properties of probability distributions, ProMP, KMP and GP allow for trajectory adaptations with via-points. On the other hand, despite GMM-GMR being formulated in terms of Gaussian distributions, the re-optimization of the learned policy requires to re-estimate the model parameters, which lie in a high-dimensional space. This makes the adaptation process very expensive, which prevents its use in unstructured environments, where the policy adjustment is essential.

Besides the generation of adaptive trajectories, another desired property in LfD is extrapolation. In this regard, there is an interesting duality between GMM-GMR and GP representations. The former covariance matrices, model the variability of the trajectories. Conversely, the latter provide a measure of the prediction uncertainty, the variance increasing with the absence of training data. This information is relevant when trying to generalize the learned motion outside of the demonstrated action space. The simultaneous exploitation of both measures is considered in KMP. Moreover, in a recent work \cite{Jaquier2019} Jaquier et al. propose a GMM-based GP for encoding the trajectory.

In the recent years, there has been a growing interest in Gaussian Processes \cite{Schulz2018}. The main advantage of GP over the previously discussed methods, is their ability to encode prior beliefs through the mean and kernel functions. This allows the representation of more complex behaviors in the regions of the action space where demonstration data is sparse. A few works have studied the use of an entirely GP-based representation in the LfD context \cite{Nguyen2008,Forte2010}. Among the most representative is the one presented by Schneider et al. \cite{Schneider2010a}. They propose a representation of a pick-and-place task that effectively encodes the task variability. Similarly, Umlauft et al. \cite{Umlauft2017} estimate the prediction uncertainty separately, using Wishart Processes. The learned trajectory is retrieved combining GP and DMP. Neither of these works consider the adaptation of the learned policy. Other works formulate the learning and motion planning problem within a single GP-based framework \cite{Rana2017, Osa2018}. In these works the entire trajectory is retrieved from an optimization perspective. However, this becomes inefficient as the length of the trajectory and the dimensionality of the learning problem increase. 

A drawback of GP is that they are usually only defined in Euclidean space, even though a formulation with non-Euclidean input space is possible in principle \cite{Lang2018}. Thus, when it comes to the modeling of task space trajectories, representation of orientation imposes great challenges, since is accompanied with additional constraints. This is an aspect disregarded in the aforementioned GP-based methods, which is critical in LfD. Some works have successfully addressed this question with DMP \cite{Zeestraten2017}, GMM-GMR \cite{Koutras2019} and KMP \cite{Huang2019c}. In a recent work, Lang et al. \cite{Lang2017} proposed an efficient representation for GP, which we have adopted.

In this work, we present a general Gaussian-Process-based learning from demonstration approach. For the purpose of clear comparison, the main contributions of the state-of-the-art and our approach are summarized in Table \ref{Table1}. We aim to unify in a single, entirely GP-based framework, the main features required for a state-of-the-art LfD approach. We show how to achieve an effective representation of the manipulation skill, inferred from the demonstrated trajectories. We unify both, the task variability and the prediction uncertainty, in a single concept we refer to as task uncertainty in the remainder of the paper. Furthermore, in order to achieve an effective generalization across demonstrations, we propose the novel Task Completion Index, for temporal alignment of task trajectories. Finally, we address the adaptation of the policy through via-points, and the modulation of the robot behavior depending on the task uncertainty through variable admittance control. The paper is structured as follows: in Section \ref{Theory} we discuss the theoretical aspects of the considered GP models; in Section \ref{Framework} we present the proposed learning from demonstration framework; in Section \ref{Application} we illustrate the main aspects of the paper through a real-world application with the TIAGo robot; finally, in Section \ref{Conclusions}, we summarize the final conclusions.

\section{Gaussian Process Models}
\label{Theory}

In this section we discuss the theoretical background of the proposed LfD approach. First, we present the fundamentals of GP \cite{Rasmussen2006}. Then, we address the challenges of modeling rigid-body dynamics with them. Finally, we present how heteroscedastic GP allows to accurately represent the uncertainty of the taught manipulation task. 

\subsection{Gaussian Process Fundamentals}

Intuitively, one can think of a Gaussian process as defining a distribution over functions, and inference taking place directly in the space of functions. Formally, GP are a collection of random variables, any finite number of which have a joint Gaussian distribution \cite{Rasmussen2006}. It can be completely specified by its mean $m(t)$ and covariance $k(t,t')$ functions:\begin{equation}
m(t)=\mathbb{E}\left[f(t)\right]\end{equation}\begin{equation}
k(t,t')=\mathbb{E}\left[\left(f(t)-m(t)\right)\left(f(t')-m(t')\right)\right]\end{equation} where $f(t)$ is the underlying process, $m(t)$ depicts the prior knowledge of its mean, and $k(t,t')$ is symmetric and positive semi-definite (usually referred to as kernel) that must be specified. We are interested in incorporating the knowledge that the training data $\mathcal{D}=\left\{\left(t_i,y_i\right)\right\}^N_{i=1}$ provides about $f(t)$. We consider that we do not have available direct observations, but only noisy versions $y$. Let $\mathbf{m}(t)$ be the vector of the mean function evaluated at all training points $t$ and $K(t,t^*)$ be the matrix of the covariances evaluated at all pairs of training and prediction points $t^*$. Assuming additive independent identically distributed Gaussian noise with variance $\sigma_n^2$, we can write the joint distribution of the observed target values $\mathbf{y}$ and the function values at the test locations $\mathbf{f}^*$ under the prior as:
\begin{equation}
\small
\left[\begin{array}{c}
     \mathbf{y} \\
     \mathbf{f}^* 
\end{array}\right]\sim \mathcal{N}\left(\left[\begin{array}{c}
     \mathbf{m}(t) \\
     \mathbf{m}(t^*) 
\end{array}\right],\left[\begin{array}{cc}
     K(t,t)+\sigma_n^2I & K(t,t^*) \\
     K(t^*,t) & K(t^*,t^*)
\end{array}\right]\right)
\end{equation} 
\normalsize

The posterior distribution over functions can be computed by conditioning the joint Gaussian prior distribution on the observations $p\left(\mathbf{f}^*|t,\mathbf{y},t^*\right)\sim \mathcal{N}\left(\boldsymbol{\mu}^*,\boldsymbol{\Sigma}^*\right)$ where:\begin{equation}
\boldsymbol{\mu}^*=\mathbf{m}(t^*)+K(t^*,t)\left[K(t,t)+\sigma_n^2I\right]^{-1}\left[\mathbf{y}-\mathbf{m}(t)\right]\end{equation}\begin{equation}\boldsymbol{\Sigma}^*=K(t^*,t^*)-K(t^*,t)\left[K(t,t)+\sigma_n^2I\right]^{-1}K(t,t^*)\end{equation} 

When we consider only the prediction of one output variable, $k(t,t')$ is a scalar function. The previous concepts can be extended to multiple-output GP (MOGP) by taking a matrix covariance function $\mathbf{k}(t,t')$. Usual approaches to MOGP modelling are mostly formulated around the Linear Model of Coregionalization (LMC) \cite{Alvarez2012}. For a $d$-dimensional outptut the kernel is expressed in the following form:\begin{equation}\small\mathbf{B}\otimes\mathbf{k}(t,t')=\left[\begin{array}{ccc}
     B_{11}k_{11}(t_1,t_1') & \hdots & B_{1d}k_{1d}(t_1,t_d')  \\
     \vdots & \ddots & \vdots \\
     B_{d1}k_{d1}(t_d,t_1') & \hdots & B_{dd}k_{dd}(t_d,t_d')
\end{array}\right]\normalsize\end{equation} where $\mathbf{B}\in\mathbb{R}^{d\times d}$ is regarded as the coregionalization matrix and $t_i$ represents the input corresponding to the $i$-th output.  Diagonal elements correspond to the single-output case, while the off-diagonal elements represent the prior assumption on the covariance of two different output dimensions \cite{Liu2018}. If no a-priori assumption is made, $B_{ij}=0$ for $i\neq j$ and the MOGP is equivalent to $d$ independent GP. 

Regarding the form of $k(t,t')$, typically kernel families have free hyperparameters $\Theta$. Such parameters can be determined by maximizing the log marginal likelihood:\normalsize\begin{equation} \log{p(\mathbf{y}|t,\Theta)}=-\frac{1}{2}\mathbf{y}^TK_y^{-1}\mathbf{y}-\frac{1}{2}\log\left|K_y\right|-\frac{N}{2}\log2\pi\end{equation}\normalsize where $K_y=K(t,t)+\sigma_n^2I$. This optimization problem might suffer from multiple local optima.

\subsection{Rigid-Body Motion Representation}
\label{Motion_Representations}

In the LfD context, representation of trajectories in task space is usually required. However, the modelling of rotations is not straightforward with GP, since the standard formulation is defined for an underlying Euclidean space. A common approach is to use the Euler angles, and exploit that locally the rotation group $SO(3) \simeq \mathbb{R}^3$, allowing distances to be computed as Euclidean. However, when this approximation is no longer valid (e.g. at low sampling frequency or if collected data is sparse) it might lead to inaccurate predictions. To overcome this issue, as proposed in \cite{Lang2017}, rotations can also be represented by a set of unit length Euler axes $\mathbf{u}$ together with a rotation angle $\theta$:\begin{equation}
SO(3)\subset \left\{\theta\mathbf{u}\in\mathbb{R}^3/\lVert \mathbf{u}\rVert=1\wedge \theta \in \left[0,\pi\right]\right\}\end{equation} 

This set defines the solid ball $B_\pi(0)$ in $\mathbb{R}^3$ with radius $0\leq r\leq \pi$ which is closed, dense and compact. Ambiguity in the representation occurs for $\theta=\pi$. To obtain an isomorphism between the rotation group $SO(3)$ and the axis-angle representation, we can fix the axis representation for $\theta=\pi$:\begin{equation}
\begin{split}
\Tilde{B}_{\pi}(0)=B_{\pi}(0)\setminus\{\pi\mathbf{u}/u_z<0\;\vee\left(u_z=0\wedge u_y<0 \right)\\ \quad\quad\quad\quad\quad\vee\left(u_z=u_y=0\wedge u_x<0 \right) \}\end{split}\end{equation} where $\mathbf{u}=\left(u_x,u_y,u_z\right)$. This parametrization is a minimal and unique $SO(3)\simeq\Tilde{B}_{\pi}(0)$. Rigid motion dynamics is given by a mapping from time, to translation and rotation $h:\mathbb{R}\longrightarrow SE(3)$. Let the translational components be defined by the Euclidean vector $\mathbf{v}\in\mathbb{R}^3$. Then $SE(3)$ is defined isomorphically by $SE(3)\simeq \mathbb{R}^3\times\Tilde{B}_{\pi}(0)$. Thus, rigid body motion can be represented in MOGP with the 6-dimensional output vector structure $\left(\mathbf{v},\theta\mathbf{u}\right)=\left(x,y,z,\theta u_x,\theta u_y, \theta u_z\right)$. 

Another possible, more accurate representation, can be achieved with dual quaternions \cite{Lang2018}. However, as shown in \cite{Lang2017}, with the proposed parametrization, a good performance is attained and computations are more efficient.

\subsection{Heteroscedastic Gaussian Process}
\label{Heteroscedastic}

\begin{figure}[h]
    \centering
    {\includegraphics[width=1.0\linewidth]{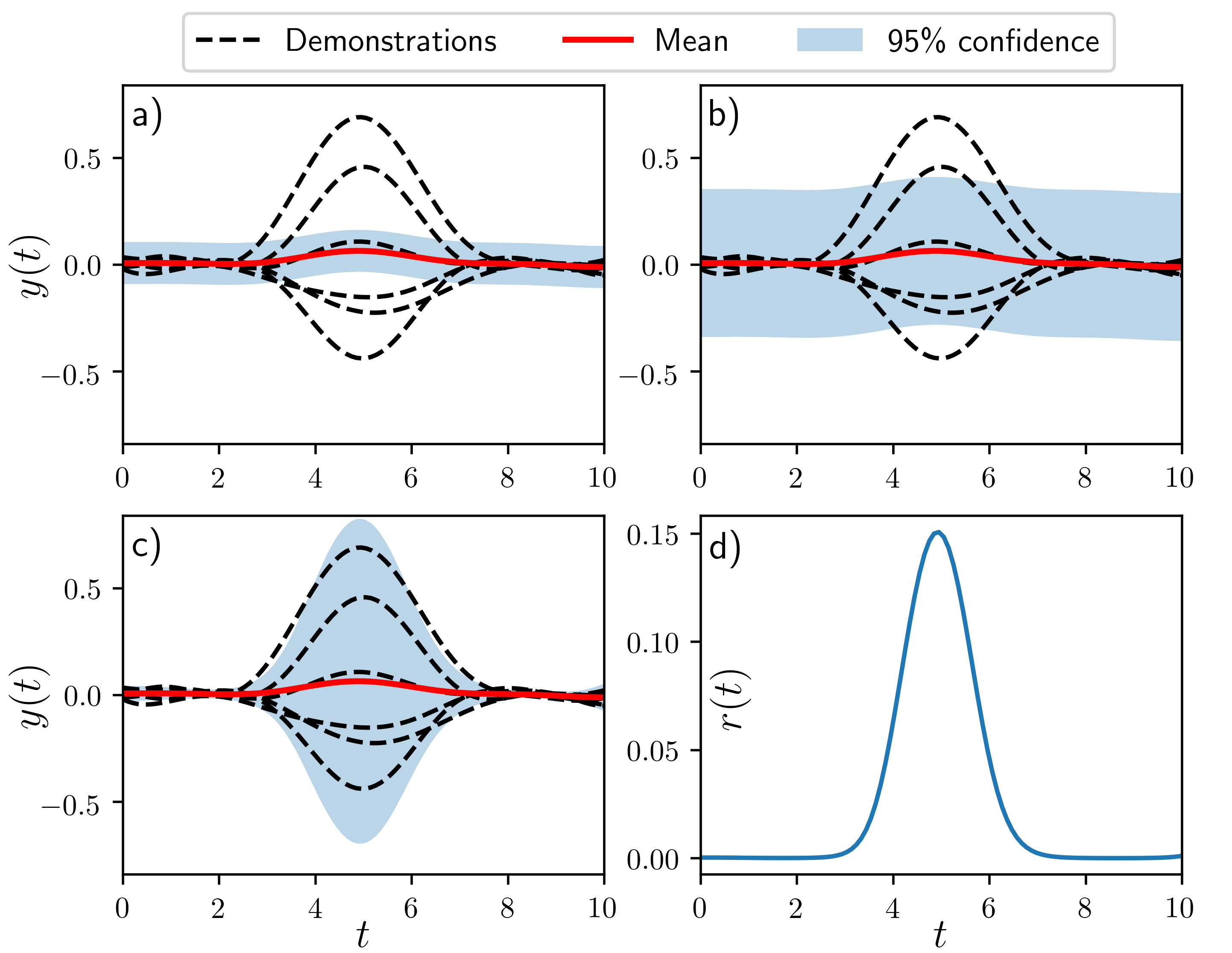}}
    \caption{Standard GP do not accurately model the uncertainty of the demonstrated task as can be seen in a) and b), where it is underestimated and overestimated, respectively. On the other hand, the heteroscedastic GP approach, in c), adequately encodes the uncertainty in the different phases of the task, considering the local noise in d).}
    \label{fig2}
    \vspace{-1.2em}
\end{figure}

The standard Gaussian Process model assumes a constant noise level. This can be an important limitation when encoding a manipulation task. Consider the example shown in Figure \ref{fig2}: it is evident that while the initial and final positions are highly constrained, that is not the case for the path to follow between such positions. In graphs a) and b) we can see that with a standard approach we accurately represent the mean but not the variability of demonstrations. 

Considering an independent normally distributed noise, $\lambda\sim\mathcal{N}\left(0,r(t)\right)$, where the variance is input-dependent and modeled by $r(t)$. The mean and covariance of the predictive distribution can be modified to \cite{Goldberg1998}: \normalsize\begin{equation}\label{het_mean}\boldsymbol{\mu}^*=\mathbf{m}(t^*)+K(t^*,t)\left[K(t,t)+R(t)\right]^{-1}\left[\mathbf{y}-\mathbf{m}(t)\right]\end{equation}\normalsize\begin{equation}\footnotesize\label{het_cov}\boldsymbol{\Sigma}^*=K(t^*,t^*)+R(t^*)-K(t^*,t)\left[K(t,t)+R(t)\right]^{-1}K(t,t^*)
\end{equation}\normalsize where $R(t)$ is a diagonal matrix, with elements $r(t)$.

Taking into account the input-dependent noise shown in Figure \ref{fig2}d) the uncertainty in the different phases of the manipulation task is effectively encoded by the uncertainty of Figure \ref{fig2}c). This approach is commonly referred to as heteroscedastic Gaussian Process. 

\section{Learning from Demonstration Framework}
\label{Framework}

In this section, we present the proposed GP-based LfD framework. First, we formalize the problem of learning manipulation skills from demonstrated trajectories. Then, we propose an approach for encoding the learned policy with GP. Next, we discuss the temporal alignment of demonstrations. We also present a method that allows to adapt the learned policy through via-points. Finally, we study how the uncertainty model of GP can be exploited to stably modulate the robot behavior, varying end-effector virtual dynamics.

\subsection{Problem Statement}

In LfD we assume that a dataset of demonstrations is available. In the trajectory-learning case, the dataset consists of a set of trajectories $\mathbf{s}$ together with a timestamp $t\in \mathbb{R}$, $\mathcal{D}=\left\{\left(t_i,\mathbf{s}_i\right)\right\}^N_{i=1}$. Without loss of generality, we will consider $\mathbf{s}_i\in SE(3)$. The aim is to learn a policy $\pi$ that infers, for a given time, the desired end-effector pose $\mathbf{s}^d_i$ to perform the taught manipulation task: $\mathbf{s}^d_i=\pi(t_i)$. The policy must generate continuous and smooth paths, and generalize over multiple demonstrations.

\subsection{Manipulation Task Representation with GP}

Representing a manipulation task using heteroscedastic GP models requires the specification of $m(t)$, $k(t,t')$ and $r(t)$. As we have discussed in Section \ref{Motion_Representations}, a suitable mapping for representing a trajectory is given by the following MOGP:\begin{equation}
\pi(t)\sim\mathcal{GP}\left(\boldsymbol{\mu}^*,\boldsymbol{\Sigma}^*\right): t\longrightarrow \left(x,y,z,\theta u_x,\theta u_y, \theta u_z\right)
\end{equation} 

The prior mean function is commonly defined as $m(t)=0$. Although not necessary in general, if no prior knowledge is available this is a simplifying assumption. The GP covariance function controls the policy function shape. The chosen kernel must generate continuous and smooth paths. Note also that the time parametrization of trajectories is invariant to translations in the time domain. Thus, the covariance function must be stationary. That is, it should be a function of $\tau=t-t'$. The Radial Basis Function (RBF) kernel fulfils all these requirements:\begin{equation}k(t,t')=\sigma_f^2\exp\left(-\frac{\left[t-t'\right]^2}{2l^2}\right)\end{equation} with hyperparameters $l$ and $\sigma_f$. Moreover, for multidimensional outputs, we have to consider the prior interaction. In the general case, we usually do not have any previous knowledge about how the different components of the demonstrated trajectories relate to each other. Thus, we can assume that the 6 components are independent a-priori. The matrix covariance function can then be written as:\begin{equation}
\mathbf{k}(t,t')=\text{diag}\left(\sigma_{f1}^2e^{\left(\left[t-t'\right]^2/l_1^2\right)},\dots,\sigma_{f6}^2e^{\left(\left[t-t'\right]^2/l_6^2\right)}\right)\end{equation} where $\text{diag}()$ refers to diagonal, and $l_i$ and $\sigma_{fi}$ correspond to output dimension $i$. In Section \ref{Heteroscedastic} we discussed the convenience of specifying an input-dependent noise function $r(t)$ for encoding the manipulation skill with GP. Usually, it is not known a-priori and must be inferred from the demonstrations. As proposed in \cite{Kersting2007}, first an standard GP can be fit to the data. Its predictions can be used to estimate the input-dependent noise empirically. Then, a second independent GP can be used to model $z(t)=\log\left[r(t)\right]$. Let $\mathcal{Z}$ be the set of noise data $\mathbf{z}=\left\{z_i\right\}_{i=1}^n$ and its predictions $\mathbf{z}^*$. The posterior predictive distribution can be approximated by:\begin{equation}\small
p\left(\mathbf{f}^*|\mathcal{D},t^*\right)=\iint p\left(\mathbf{f}^*|\mathcal{D},\mathcal{Z},t^*\right)p\left(\mathcal{Z}|\mathcal{D},t^*\right)\simeq p\left(\mathbf{f}^*|\mathcal{D},\mathcal{Z},t^*\right)\end{equation}\normalsize where $\mathcal{Z}=\argmax_{\mathbf{z},\mathbf{z}^*}p\left(\mathbf{z},\mathbf{z}^*|\mathcal{D},t^*\right)$. Therefore, we have specified all the required functions of the model.

\subsection{Temporal Alignment of Demonstrations}

\begin{figure}[t]
    \centering
    {\includegraphics[width=0.97\linewidth]{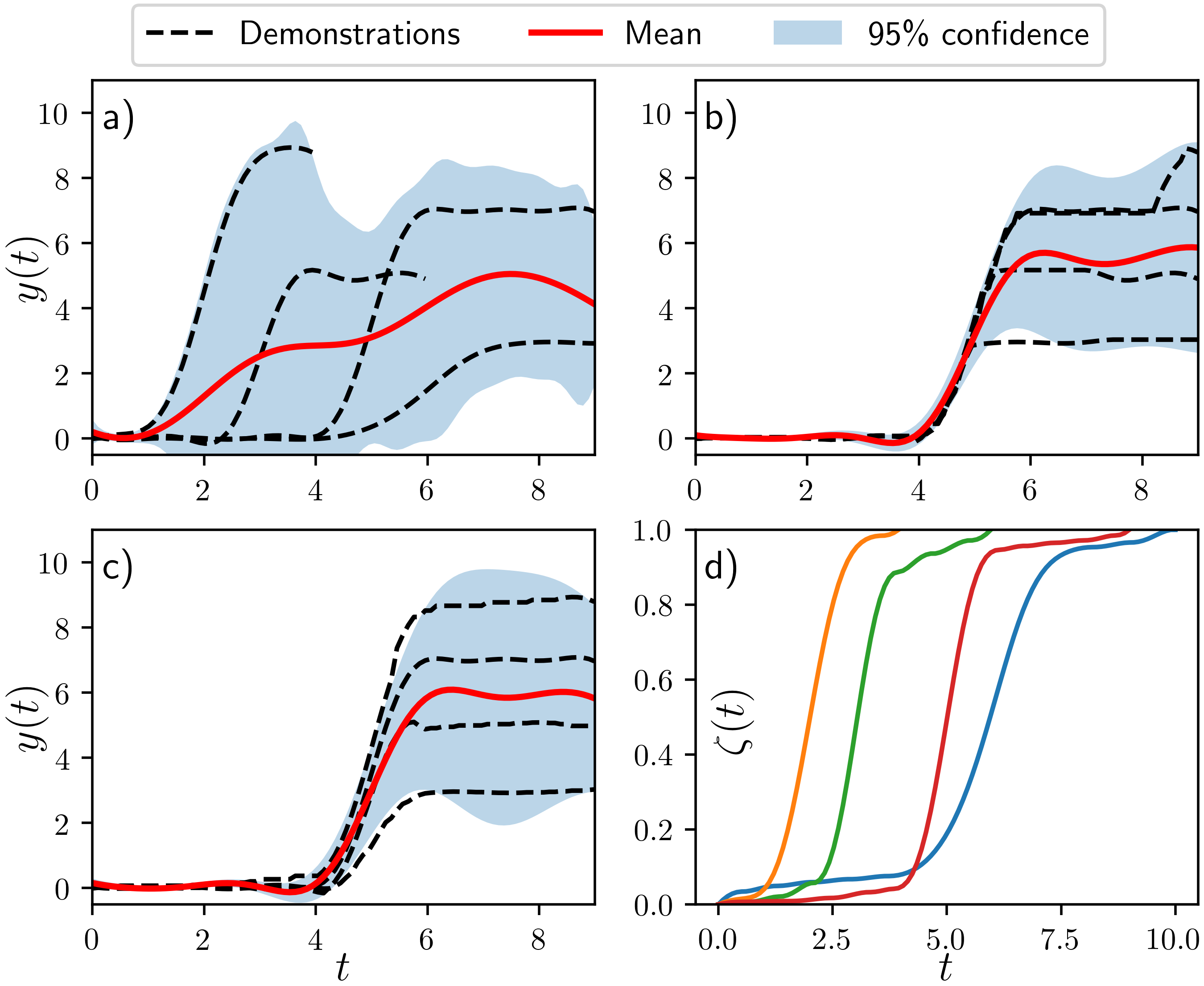}}
    \caption{In a) we observe that due to distortion in time, task constraints are not encoded correctly. In b) the trajectories are aligned with DTW using the Euclidean distance as similarity measure. In c) we show the resulting alignment using the proposed TCI d), as similarity measure.}
    \vspace{-1.2em}
    \label{fig3}
\end{figure}

For inferring a time dependent policy, the correlation between the temporal and spatial coordinates of two demonstrations of the same task must remain constant. In general, it is very difficult for a human to repeat them at the same velocity. Thus, a time distortion appears (Figure \ref{fig3}a), and should be adequately corrected. Dynamic Time Warping (DTW) \cite{Senin2008} is a well-known algorithm for finding the optimal match between two temporal sequences, which may vary in speed. 

The algorithm finds a non-linear mapping of the demonstrated trajectories and a reference based on a similarity measure. A common measure in the LfD context is the Euclidean distance. This relies on the assumption that the manipulation task can be performed always following the same path. For instance, consider the case of a pick-and-place task where the objects have to be placed in shelves at different levels. Using the Euclidean distance as similarity measure will lead to an erroneous temporal alignment (Figure \ref{fig3}b), since intermediate points for placing the object at a higher level can be mapped to ending points of a lower level. We propose to use an index which considers the portion of the trajectory that has been covered for task completion as a similarity measure. We will refer to it as the Task Completion Index (TCI). We define it in discrete form as:\begin{equation}\zeta(t_k)=\frac{\sum_{j=1}^{k}d(\mathbf{s}_{j},\mathbf{s}_{j-1})}{\sum_{j=1}^{M}d(\mathbf{s}_{j},\mathbf{s}_{j-1})}\quad\quad\forall\ k=1,\dots M\end{equation} where $\mathbf{s}_{j}\in SE(3)$ refers to the trajectory point at time instant $t_j$, $d(,)$ to an scalar distance function and $M$ to the total number of discrete points. Note that $0=\zeta(t_0)\leq\zeta(t_k)\leq\zeta(t_M)=1$. As a distance function on $SE(3)$, using the representation discussed in Section \ref{Motion_Representations}, we define:\begin{equation}d(\mathbf{s}_{i},\mathbf{s}_{j})=\sqrt{\omega_1\left[d_{arc}(\theta_{i}\mathbf{u}_{i},\theta_{j}\mathbf{u}_{j})\right]^2+\omega_2\lVert\mathbf{v}_{i}-\mathbf{v}_{j}\rVert^2}\end{equation} where $\omega_k$ are a convex combination of weights for application dependent scaling and $d_{arc}(,)$ is the length of the geodesic between rotations \cite{Lang2017}:
\begin{equation}
\footnotesize
d_{arc}(\theta_{i}\mathbf{u}_{i},\theta_{j}\mathbf{u}_{j})=2\arccos\left|\cos\frac{\theta_{i}}{2}\cos\frac{\theta_{j}}{2}+\sin\frac{\theta_{i}}{2}\sin\frac{\theta_{j}}{2}\mathbf{u}_i^T\mathbf{u}_j\right|\end{equation}

In Figure \ref{fig3}c we show that the trajectories are warped correctly, allowing then an effective encoding of the manipulation task, with the proposed TCI (Figure \ref{fig3}d).
 
\subsection{Policy Adaptation through Via-points}

The modulation of the learned policy through via-points is an important property to adapt to new situations. Let $\mathcal{V}=\left\{\left(t_i,\mathbf{s}_i^v\right)\right\}$ be the set of via-points $\mathbf{s}_i^v$ which are desired to be reached by the policy at time instant $t_i$. In the proposed probabilistic framework, generalization can be implemented by conditioning the policy on both $\mathcal{D}$ and $\mathcal{V}$. Assuming that the predictive distribution of each set can be computed independently, the conditioned policy is \cite{Deisenroth2015a}:\begin{equation}
p\left(\mathbf{f}^*|\mathcal{D},\mathcal{V},t^*\right)=p\left(\mathbf{f}^*|\mathcal{D},t^*\right)p\left(\mathbf{f}^*|\mathcal{V},t^*\right)\end{equation} 

If $p\left(\mathbf{f}^*|\mathcal{D},t^*\right)\sim\mathcal{N}\left(\mu^d,\Sigma^d\right)$ and $p\left(\mathbf{f}^*|\mathcal{V},t^*\right)\sim\mathcal{N}\left(\mu^v,\Sigma^v\right)$, then, it holds that $ p\left(\mathbf{f}^*|\mathcal{D},\mathcal{V},t^*\right)\sim\mathcal{N}\left(\mu^{**},\Sigma^{**}\right)$ where:\begin{equation}
\mu^{**}=\Sigma^v\left(\Sigma^d+\Sigma^v\right)^{-1}\mu^d+\Sigma^d\left(\Sigma^d+\Sigma^v\right)^{-1}\mu^v\end{equation}\begin{equation}
\Sigma^{**}=\Sigma^d\left(\Sigma^d+\Sigma^v\right)^{-1}\Sigma^v\end{equation} 

\begin{figure}[h]
    \centering
    {\includegraphics[width=1.0\linewidth]{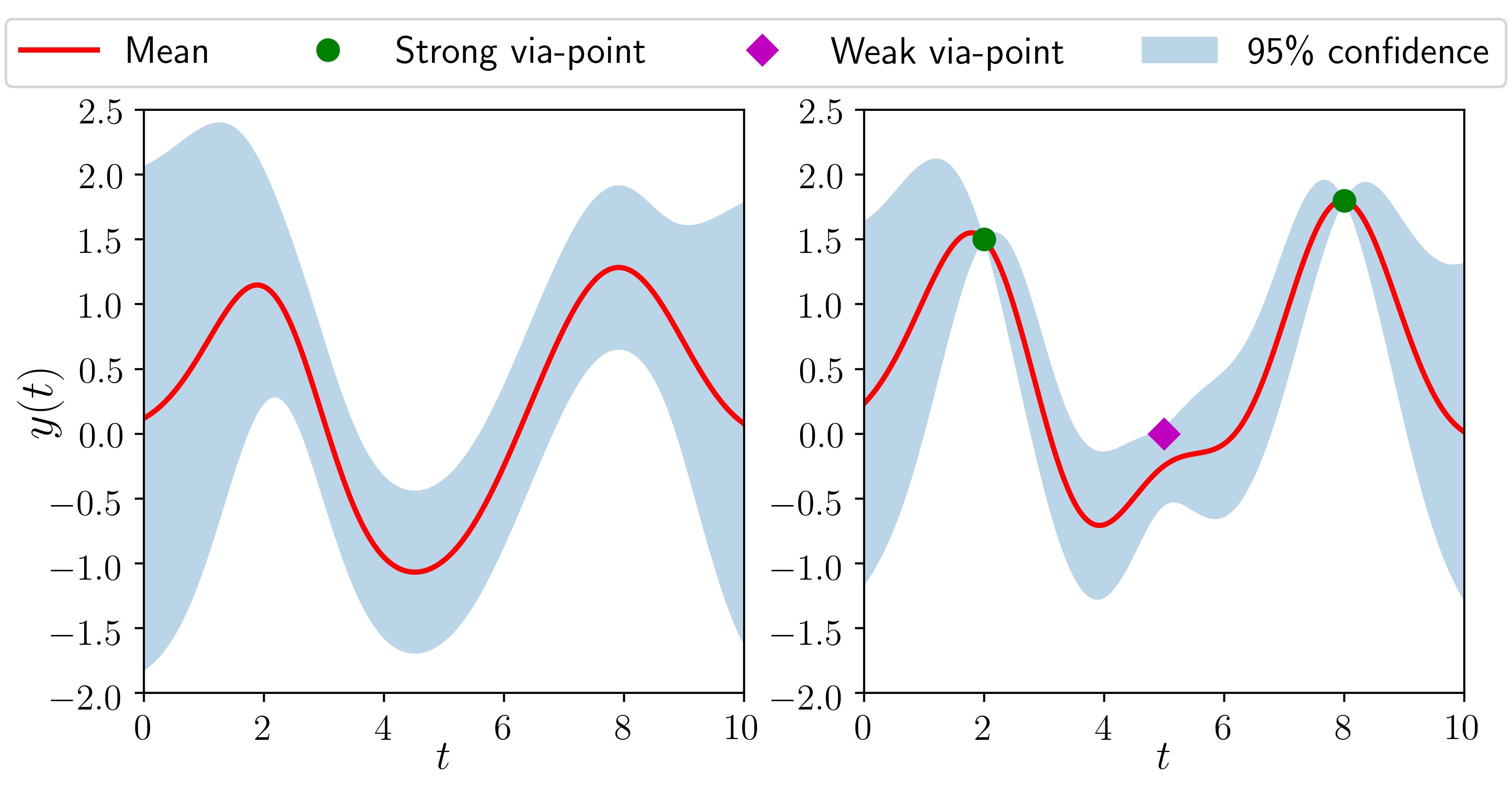}}
    \caption{On the left, a GP model based on the demonstrated trajectories. On the right, the policy adapted through via-points.}
    \label{fig4}
    \vspace{-1.1em}
\end{figure}

The resulting distribution is computed as a product of Gaussians, and is a compromise between the via-point constraints and the demonstrated trajectories, weighted inversely by their variances. Considering an heteroscedastic GP model for $\mathcal{V}$ (equations \ref{het_mean} and \ref{het_cov}), the strength of the via-point constraints can then be easily specified by means of the latent noise function. For instance, via-points with low noise will have a higher relative weight, modifying significantly the learned policy. On the other hand, via-points with a high noise level will produce a more subtle effect. In Figure \ref{fig4} we illustrate how the distribution adapts to strong and weak defined via-points. 

It should be remarked that the posterior predictive distribution of $\mathcal{D}$ only needs to be computed once. Thus, adaptation of the policy just involves a computational cost of $\mathcal{O}\left(m^3\right)$, where $m$ is the number of predicted outputs. Since $m$ can be specified, the proposed approach is suitable for on-line applications (for further insight on GP complexity see \cite{Bilj2018}).

\subsection{Modulation of the Robot Behavior}
\label{Admittance}

\setcounter{figure}{5}
\begin{figure*}[t]
    \centering
    {\includegraphics[width=1.0\linewidth]{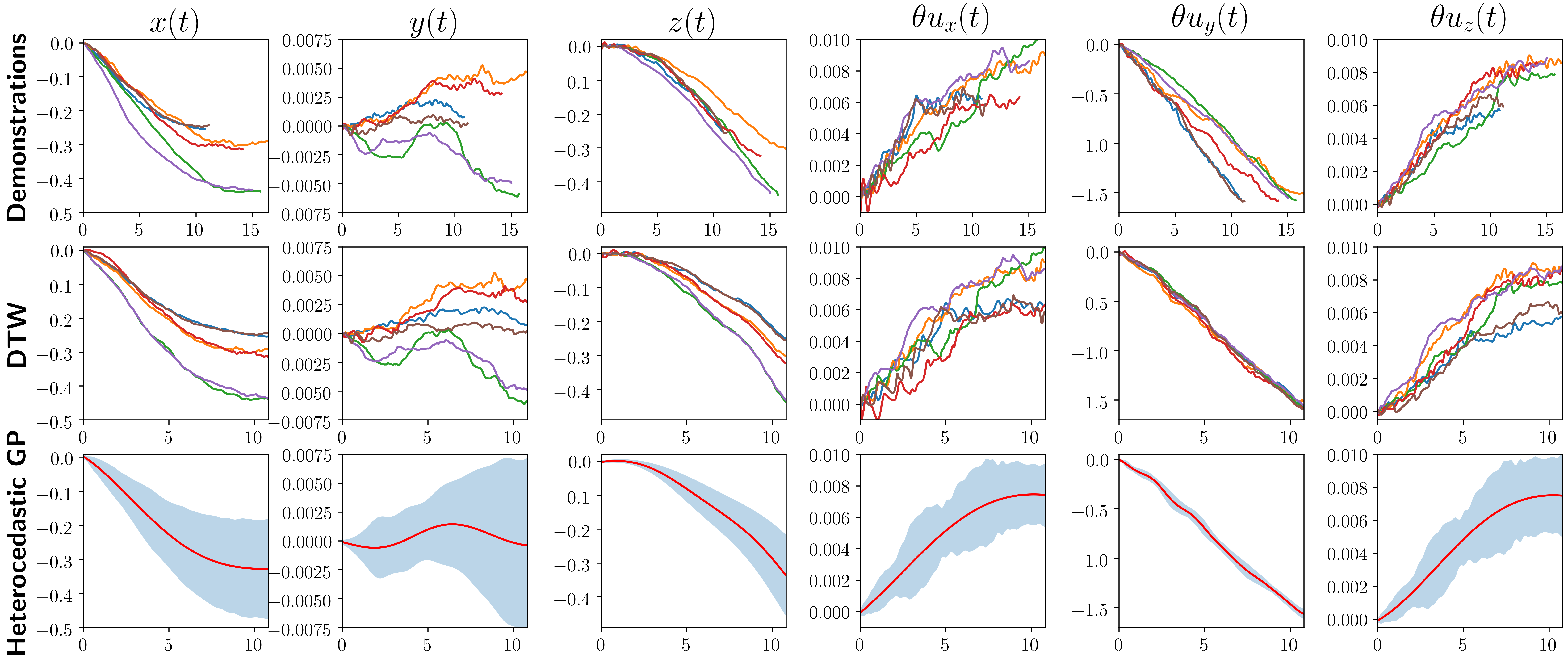}}
    \caption{In the first row, demonstrated trajectories. In the second row, alignment with DTW using TCI index. In the third row, heteroscedastic GP of each dimension of the MOGP encoding the learned policy. Note that significant variations are only observed in $x$, $z$ and $\theta u_y$ (see vertical axis scale).}
    \label{fig6}
    \vspace{-1.0em}
\end{figure*}

In LfD, it is often convenient to adapt the behavior of the robot as a function of the uncertainty in the different phases of the task. Let the robot end-effector be controlled through a virtual spring-mass-damper model dynamics:\begin{equation}
\mathbf{M}\left(t\right)\ddt{\mathbf{e}}\left(t\right)+\mathbf{D}\left(t\right)\dt{\mathbf{e}}\left(t\right)+\mathbf{K}_p\left(t\right)\mathbf{e}\left(t\right)=\mathbf{F_{ext}}\left(t\right)\end{equation} where $\mathbf{M}\left(t\right),\mathbf{D}\left(t\right),\mathbf{K}_p\left(t\right)\in\mathbb{R}^{6\times 6}$ refer to inertia, damping and stiffness, respectively, and $\mathbf{e}\left(t\right)\in\mathbb{R}^{6\times 1}$ is the tracking error, when subjected to an external force $\mathbf{F_{ext}}\left(t\right)\in\mathbb{R}^{6\times 1}$. 

It can be proved (see \cite{Kronander2016}) that for a constant, symmetric, positive definite $\mathbf{M}$, and $\mathbf{D}\left(t\right)$, $\mathbf{K}_p\left(t\right)$ continuously differentiable, the system is globally asymptotically stable if there exists a $\gamma>0$ such that:

\begin{enumerate}
    \item $\gamma\,\mathbf{M}-\mathbf{D}\left(t\right)$ is negative semidefinite
    \item $\dt{\mathbf{K}}_p\left(t\right)+\gamma\,\dt{\mathbf{D}}\left(t\right)-2\gamma\,\mathbf{K}_p\left(t\right)$ is negative definite
\end{enumerate}

Now consider $\mathbf{M}$, $\mathbf{D}\left(t\right)$ and $\mathbf{K}_p\left(t\right)$ diagonal matrices, and a constant damping ratio $\delta$. Substituting $d\left(t\right)=2\delta\sqrt{m\,k_p\left(t\right)}$ on the second stability condition, it yields the following upper bound for the stiffness derivative:\begin{equation}
\label{eq23}
\dt{k}_p\left(t\right)<\frac{2\gamma\sqrt{k_p\left(t\right)^3}}{\sqrt{k_p\left(t\right)}+2\delta\,\gamma\sqrt{m}}\end{equation} where $m$ and $k_p(t)$ are an arbitrary diagonal element of $\mathbf{M}$ and $\mathbf{K}_p\left(t\right)$, respectively. In order to modulate the robot behavior, we propose the following variable stiffness profile:\begin{equation}
\label{eq24}
k_p(t)=k_p^{max}-\frac{k_p^{max}-k_p^{min}}{1+e^{-\alpha\left(\sigma(t)-\beta\right)}}\end{equation} which increases the stiffness inversely to the uncertainty $\sigma(t)$ and saturates at $k_p^{min}$ and $k_p^{max}$ for high and low values respectively. Differentiating we have:\begin{equation}
\label{eq25}
\dt{k}_p(t)=\alpha k_p(t)\left(1-\frac{k_p(t)}{k_p^{max}-k_p^{min}}\right)\frac{d\sigma(t)}{dt}\end{equation} 

For a constant $d\sigma(t)/dt$, the maximum value of the stiffness derivative $\dt{k}_p(t)$ is obtained for $k_p(t)=\left(k_p^{max}-k_p^{min}\right)/2$. Thus, substituting in $\left(\ref{eq25}\right)$, it yields the following upper bound:\begin{equation}
\label{eq26}
    \dt{k}_p(t)\leq\frac{\alpha}{4}\left(k_p^{max}-k_p^{min}\right)\frac{d\sigma(t)}{dt}
\end{equation}

Then, from inspection of the first stability condition, we can see that $\gamma$ defines a lower bound for the minimum allowed damping $d(t)$. Thus, given the variable stiffness profile in Equation \ref{eq24}, and assuming constant damping ratio, the most restrictive value is $\gamma=2\delta\sqrt{k_p^{min}/m}$. Substituting in $\left(\ref{eq23}\right)$, we can obtain the following lower bound:
\begin{equation}
\label{eq27}
    \dt{k}_p(t)<\frac{4\delta\sqrt{\left(k_p^{min}\right)^3}}{\left(1+4\delta^2\right)\sqrt{m}} \leq  \frac{2\gamma\sqrt{k_p\left(t\right)^3}}{\sqrt{k_p\left(t\right)}+2\delta\,\gamma\sqrt{m}}
\end{equation}

Then, from equations $\left(\ref{eq26}\right)$ and $\left(\ref{eq27}\right)$ the following sufficient stability condition can be derived:\begin{equation}\frac{d\sigma(t)}{dt}<\frac{16\delta}{\alpha}\frac{\sqrt{\left(k_p^{min}\right)^3}}{\left(k_p^{max}-k_p^{min}\right)\left(1+4\delta^2\right)\sqrt{m}}\end{equation} 

The control parameters can then be tuned to ensure the satisfaction of this inequality. Note that sharper uncertainty profiles $\sigma(t)$ are more restrictive with respect to variations of the stiffness. For instance, stability is favored by a smaller range $\left(k_p^{max}-k_p^{min}\right)$ or lower values of $\alpha$, i.e. slower transition between stiff and compliant behaviors. For the limit cases $k_p^{max}\longrightarrow k_p^{min}$ and $\alpha\longrightarrow 0$, that is, constant stiffness, stability can be achieved regardless of $\sigma(t)$. It can also be observed, since the right-hand side of the inequality is always positive, that with the proposed variable stiffness profile, stability is ensured if the uncertainty decreases.

\section{An Example Application: Door Opening Task}
\label{Application}

In order to illustrate how the proposed GP-based LfD approach can be applied to real-world manipulation tasks, we address the problem of opening doors using a TIAGo robot. This is a relevant skill for robots operating in domestic environments \cite{Kim2004}, since they need to open doors when navigating, to pick up objects in fetch-and-carry applications or assist people in their mobility.

\vspace{-0.5mm}
\subsection{Policy Inference from Human Demonstrations}

We performed human demonstrations using an Xsens MVN motion capture system. Right hand trajectories of the human teacher relative to the initial closed door position were recorded for three different doors (Figure \ref{fig5}). Coordinate axes were chosen such as the pulling direction is parallel to the $x$ axis and the $y$ axis is perpendicular to the floor. The demonstration dataset consisted in a total of 6 trajectories, two per each door.

\setcounter{figure}{4}
\begin{figure}[h]
    \centering
    {\includegraphics[width=0.75\linewidth]{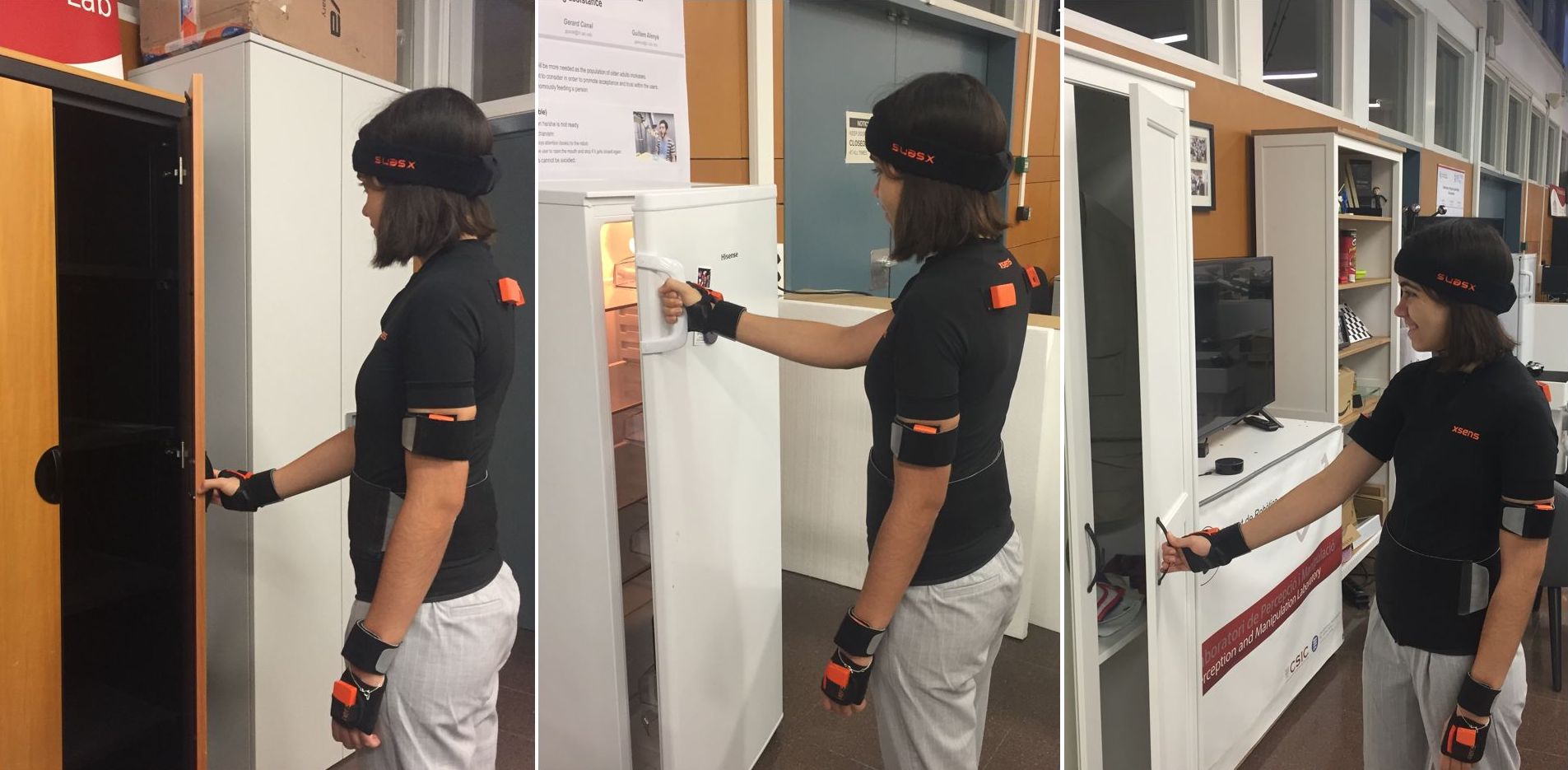}}
    \caption{Demonstrations were recorded using an Xsens MVN motion capture system. The teacher opens three doors with different radius.}
    \label{fig5}
\end{figure}

\vspace{-1mm}
The recorded trajectories were then temporally aligned using DTW and the proposed TCI index. Next, the data was used to infer the door opening policy. In Figure \ref{fig6} we show these steps for each output dimension. On the third row, we can see the resulting heteroscedastic MOGP representation. Note that in the door opening motion, significant variations are only observed in the $x$, $z$ and $\theta u_y$ components. We can see that the trajectories are warped effectively using the proposed TCI similarity measure, since they are clearly clustered in three groups, one for each type of door. It can also be observed that the resulting heteroscedastic MOGP effectively encodes the skill. This is more evident in Figure \ref{fig7}, where position uncertainty has been projected onto the $x-z$ plane. We can observe the uncertainty in the door radius is accurately captured from demonstrations. 

\setcounter{figure}{8}
\begin{figure*}[h]
    \centering
    {\includegraphics[width=1.0\linewidth]{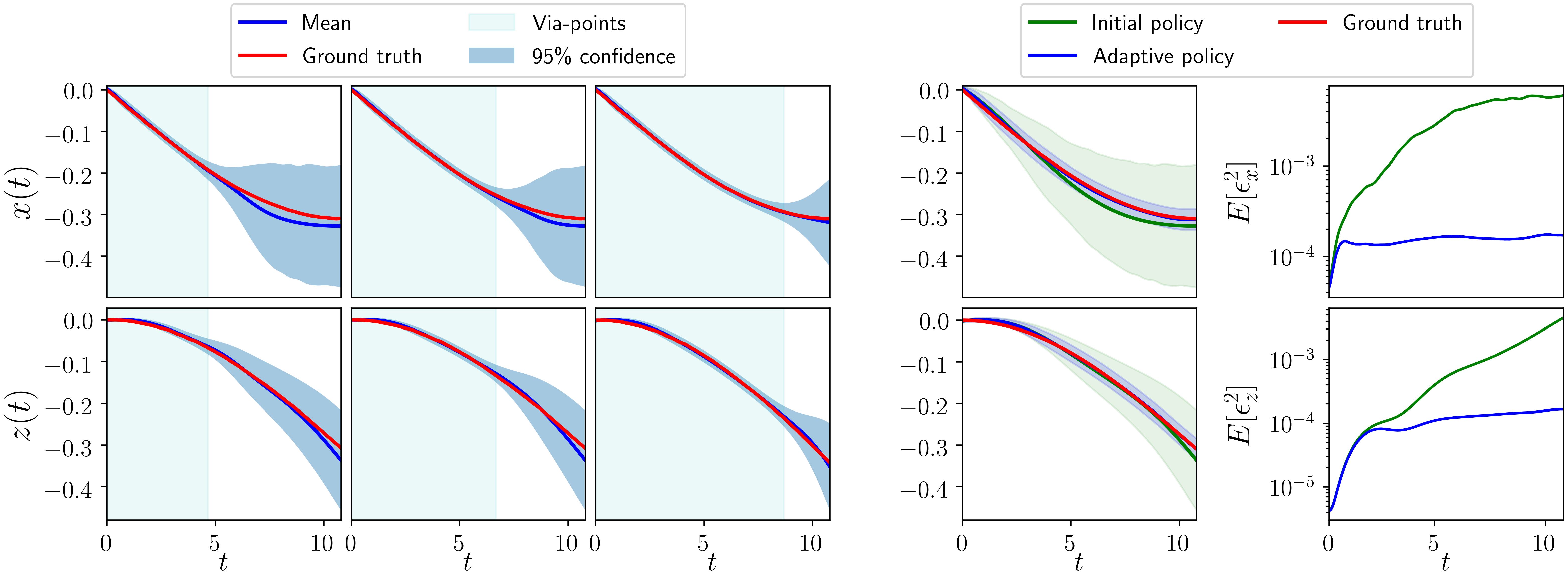}}
    \caption{The first three columns show how does the posterior distribution vary considering as via-points the observations of the door motion in the light-blue shaded area. The next column shows the comparison between the predictive distribution considering the adaptive policy or the policy based only on human demonstrations. The shaded areas and lines of the same color correspond to the 95\% confidence interval and mean, respectively. The column on the right shows the mean squared prediction error of each policy.}
    \label{fig9}
    \vspace{-1.2em}
\end{figure*}

\setcounter{figure}{6}
\begin{figure}[h]
    \centering
    {\includegraphics[width=0.67\linewidth]{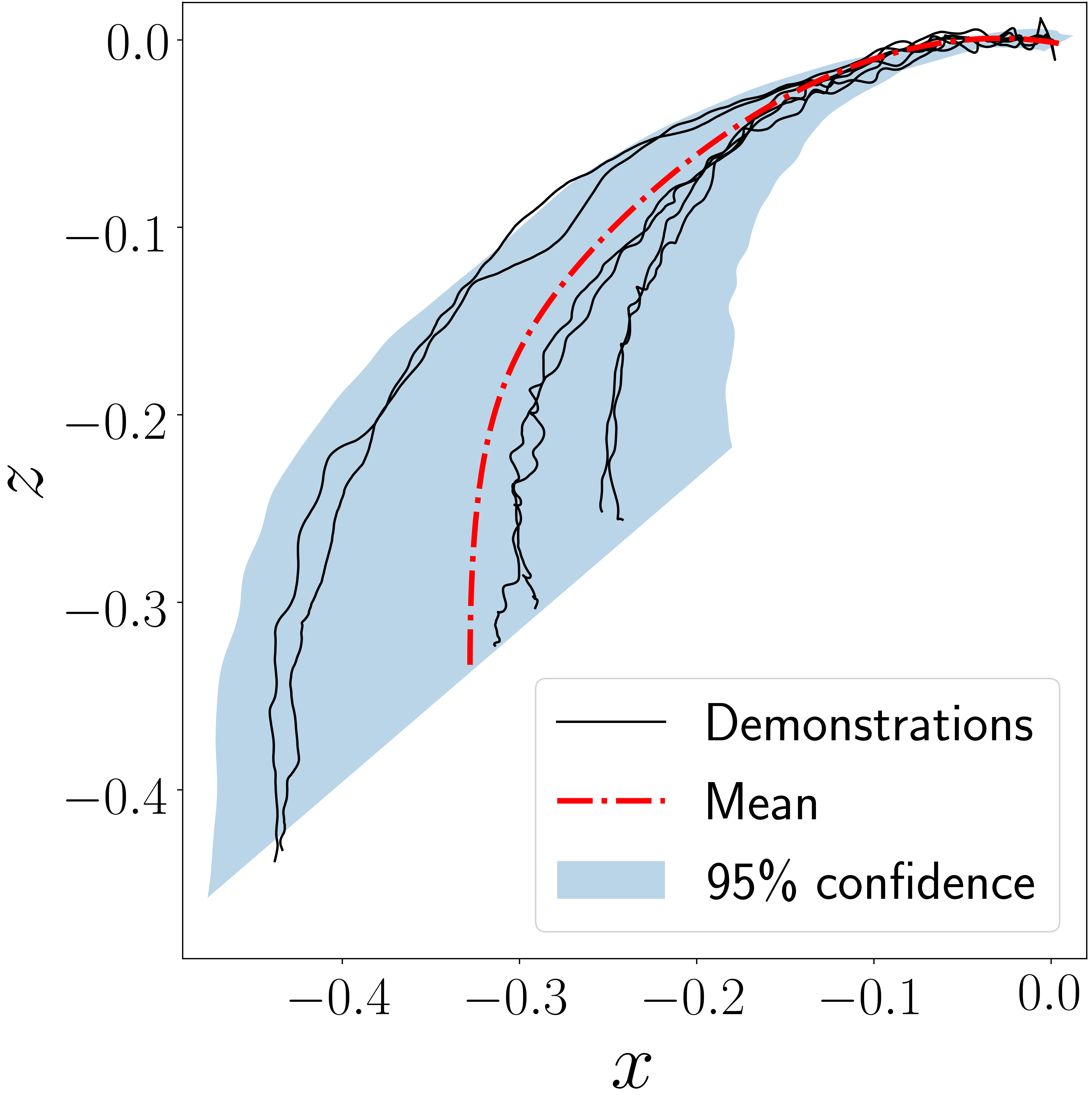}}
    \caption{Door opening policy projected on the $x-z$ plane.}
    \label{fig7}
    \vspace{-0.5em}
\end{figure}

\subsection{Policy Adaptation and Modulation of the Robot Behavior}

In the reproduction stage, we gather observations of the door motion solving the forward kinematics of the robot. These observations can then be defined as via-points to improve the policy prediction capabilities. Additionally, we can take advantage of the forces exerted by the door to correct small biases on the policy, adopting a variable admittance control scheme. The set-point of the controller is changed through the virtual dynamics discussed in Section \ref{Admittance}. 

\setcounter{figure}{7}
\begin{figure}[h]
    \centering
    {\includegraphics[width=0.86\linewidth]{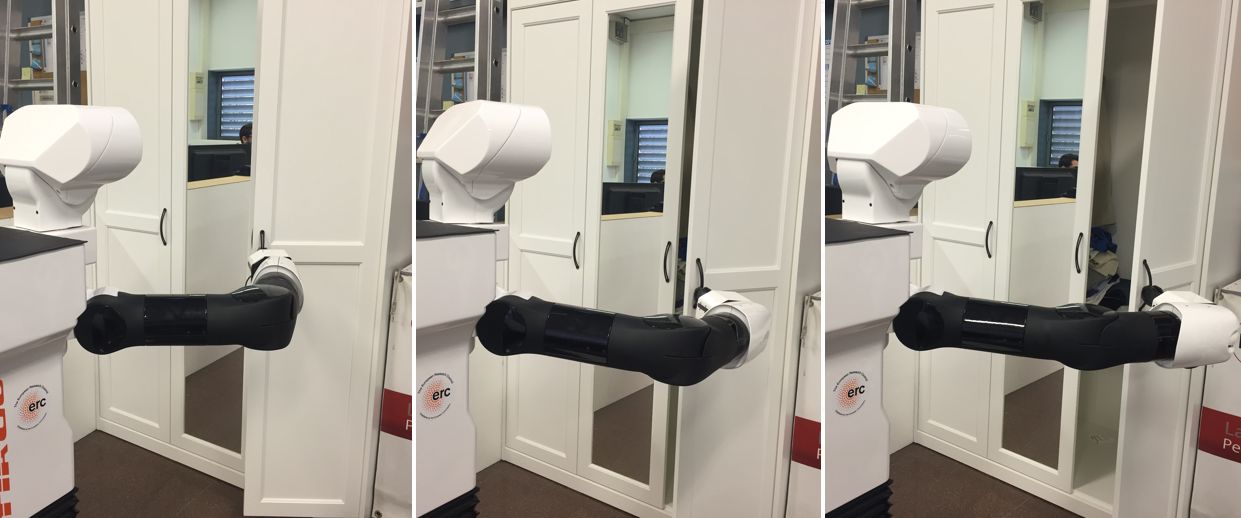}}
    \caption{TIAGo robot opening the door.}
    \label{fig8}
\end{figure}

With the proposed approach we successfully performed the door opening task with TIAGo (Figure \ref{fig8}). We can see in Figure \ref{fig9}, on the first three columns on the left, how the posterior distribution varies for coordinates $x$ and $z$ as the door is opened. We show that the adaptive policy converges to the ground truth trajectory. On the next column, we show the comparison between the resulting distribution at step $t_i$ considering via-points up to $t_{i-1}$, and the initial policy. In order to obtain a quantitative measure of the prediction performance, we evaluated the mean squared prediction error, $E\left[\epsilon^2\right]=\left(E\left[f^*(t^*)]-f(t^*)\right]\right)^2+\text{var}\left[f^*(t^*)\right]$. We can observe that the adaptive policy clearly achieves a better performance. 

The resulting variable stiffness profile is shown in Figure \ref{fig10}. We have tuned the parameters empirically, being the used values $k_p^{max}=500$, $k_p^{min}=100$, $m=1$, $\delta=1$, $\alpha=600$ and $\beta=0.01$. For simplicity, we have considered the same law for the 6 degrees of freedom. We can observe that the robot behavior is modulated towards a more compliant behavior towards the final phases, where the policy is more uncertain. We can also see that the stability bound is not crossed, which is coherent with the behavior observed in the conducted experiments, where no instabilities occurred.

\vspace{-2mm}

\setcounter{figure}{9}
\begin{figure}[h]
    \centering
    {\includegraphics[width=0.95\linewidth]{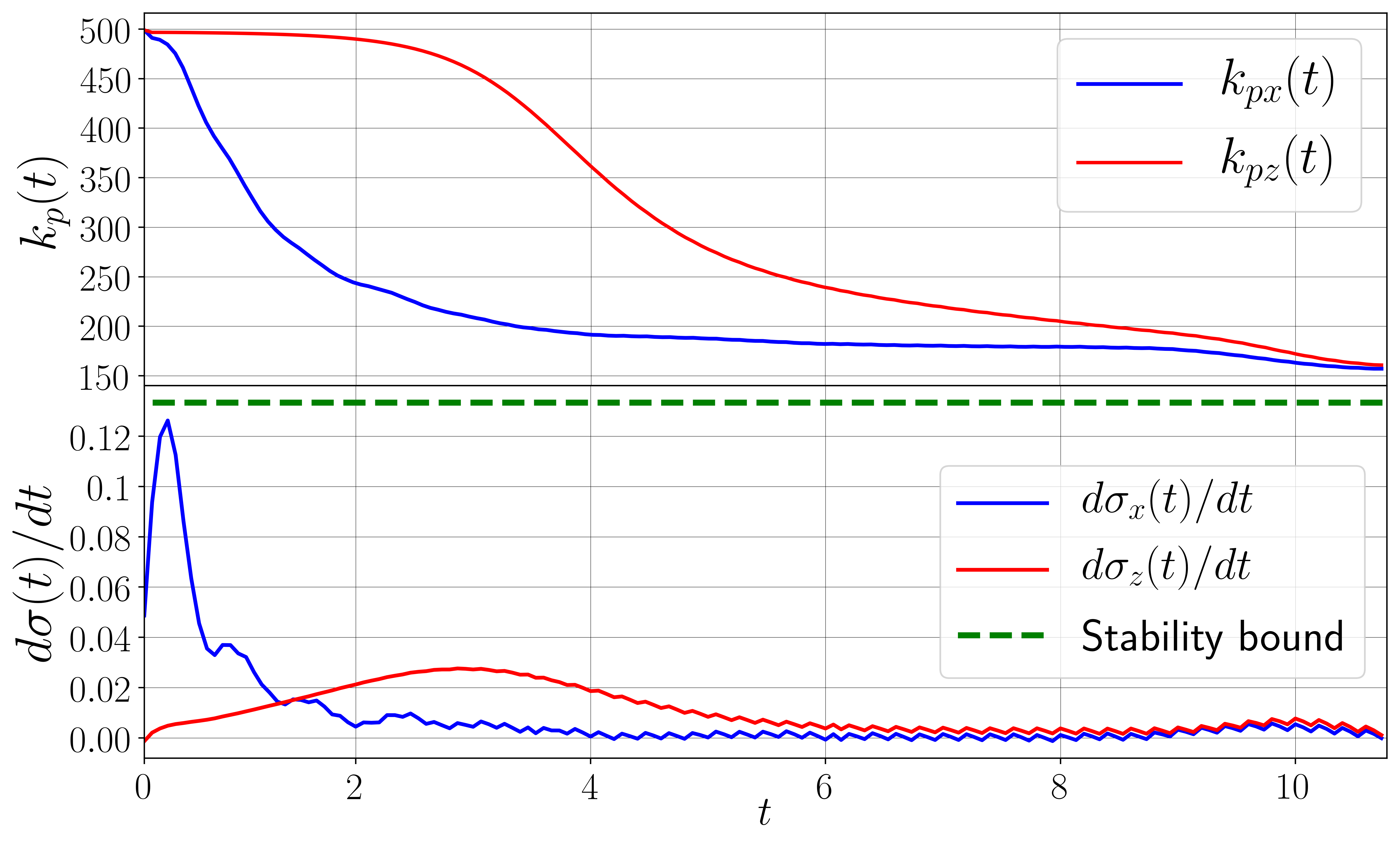}}
    \caption{On top, variable stiffness profile for $x$ and $z$. Below, the evolution of the uncertainty derivative of the adaptive policy.}
    \label{fig10}
\end{figure}

\section{Conclusion}
\label{Conclusions}

Gaussian Processes (GP) are a promising paradigm for learning manipulation skills from human demonstrations. In this paper, we present a novel approach that takes advantage of the versatility and expressiveness of these models to encode task policies. We propose an heteroscedastic multi-output GP policy representation, inferred from demonstrations. This model considers a suitable parametrization of task space rotations for GP and ensures that only continuous and smooth paths are generated. Furthermore, the introduction of an input-dependent latent noise function allows an effective simultaneous encoding of the prediction uncertainty and the variability of demonstrated trajectories.

In order to effectively establish a correlation between temporal and spatial coordinates, demonstrations must be aligned. This operation can be performed with the Dynamic Time Warping algorithm. We introduce the novel Task Completion Index, a similarity measure that allows to achieve an effective warping when the learned task requires the consideration of different paths. Adaptation of the policy can be performed by conditioning it on a set of specified via-points. We also introduce a new computationally efficient method, where the relative importance of the constraints can also be defined. Additionally, we propose an innovative variable stiffness profile that takes advantage of the uncertainty measure provided by the GP model to stably modulate the robot end-effector dynamics. 

We illustrated the proposed learning from demonstration framework through the door opening task and evaluated the performance of the learned policy through real-world experiments with the TIAGo robot. Results show that the manipulation skill is effectively encoded and a successful reproduction can be achieved taking advantage of the presented policy adaptation and robot behavior modulation approaches. 

This work aims to push the state-of-the-art in learning from demonstration towards easily extending robot capabilities. Future research will be conducted focusing on its applicability on complex tasks, such as cloth manipulation.


\bibliographystyle{IEEEtran}
\bibliography{Referencias}

\begin{thebibliography}{10}
\providecommand{\url}[1]{#1}
\csname url@samestyle\endcsname
\providecommand{\newblock}{\relax}
\providecommand{\bibinfo}[2]{#2}
\providecommand{\BIBentrySTDinterwordspacing}{\spaceskip=0pt\relax}
\providecommand{\BIBentryALTinterwordstretchfactor}{4}
\providecommand{\BIBentryALTinterwordspacing}{\spaceskip=\fontdimen2\font plus
\BIBentryALTinterwordstretchfactor\fontdimen3\font minus
  \fontdimen4\font\relax}
\providecommand{\BIBforeignlanguage}[2]{{%
\expandafter\ifx\csname l@#1\endcsname\relax
\typeout{** WARNING: IEEEtran.bst: No hyphenation pattern has been}%
\typeout{** loaded for the language `#1'. Using the pattern for}%
\typeout{** the default language instead.}%
\else
\language=\csname l@#1\endcsname
\fi
#2}}
\providecommand{\BIBdecl}{\relax}
\BIBdecl

\bibitem{Ravichandar2020}
H.~Ravichandar, A.~Polydoros, S.~S.~Chernova, and A.~Billard, ``Recent advances
  in robot learning from demonstration,'' \emph{Annual Review of Control,
  Robotics, and Autonomous Systems}, vol.~3, May 2020.

\bibitem{Colome2020}
A.~Colom\'e and C.~Torras, \emph{Reinforcement Learning of Bimanual Robot
  Skills}.\hskip 1em plus 0.5em minus 0.4em\relax Springer Tracts in Advanced
  Robotics (STAR), vol. 134. Springer International Publishing, 2020.

\bibitem{Ijspeert2001}
A.~J. {Ijspeert}, J.~{Nakanishi}, and S.~{Schaal}, ``Trajectory formation for
  imitation with nonlinear dynamical systems,'' in \emph{Proceedings 2001
  IEEE/RSJ International Conference on Intelligent Robots and Systems}, vol.~2,
  Oct 2001, pp. 752--757.

\bibitem{Pastor2009}
P.~{Pastor}, H.~{Hoffmann}, T.~{Asfour}, and S.~{Schaal}, ``Learning and
  generalization of motor skills by learning from demonstration,'' in
  \emph{IEEE International Conference on Robotics and Automation}, May 2009,
  pp. 763--768.

\bibitem{Paraschos2018}
A.~Paraschos, C.~Daniel, J.~Peters, and G.~Neumann, ``Using probabilistic
  movement primitives in robotics,'' \emph{Autonomous Robots}, vol.~42, no.~3,
  pp. 529--551, March 2018.

\bibitem{Calinon2016}
S.~Calinon, ``A tutorial on task-parameterized movement learning and
  retrieval,'' \emph{Intelligent Service Robotics}, vol.~9, no.~1, Jan 2016.

\bibitem{Pignat2019}
E.~Pignat and S.~Calinon, ``{B}ayesian {G}aussian {M}ixture {M}odel for robotic
  policy imitation,'' \emph{{IEEE} Robotics and Automation Letters ({RA-L})},
  vol.~4, no.~4, pp. 4452--4458, 2019.

\bibitem{Huang2019b}
Y.~{Huang}, L.~{Rozo}, J.~{Silv\'erio}, and D.-G. {Caldwell}, ``Non-parametric
  imitation learning of robot motor skills,'' in \emph{IEEE International
  Conference on Robotics and Automation (ICRA)}, 2019, pp. 5266--5272.

\bibitem{Huang2019c}
Y.~{Huang}, F.~J. {Abu-Dakka}, J.~{Silv\'erio}, and D.-G. {Caldwell},
  ``Generalized orientation learning in robot task space,'' in \emph{IEEE
  International Conference on Robotics and Automation (ICRA)}, 2019.

\bibitem{Nguyen2008}
D.~{Nguyen-Tuong} and J.~{Peters}, ``Local {G}aussian {P}rocess regression for
  real-time model-based robot control,'' in \emph{IEEE/RSJ International
  Conference on Intelligent Robots and Systems}, Sept 2008.

\bibitem{Forte2010}
D.~{Forte}, A.~{Ude}, and A.~{Kos}, ``Robot learning by {G}aussian {P}rocess
  regression,'' in \emph{19th International Workshop on Robotics in
  Alpe-Adria-Danube Region (RAAD 2010)}, June 2010, pp. 303--308.

\bibitem{Silverio2018}
J.~{Silv\'erio}, Y.~{Huang}, L.~{Rozo}, S.~{Calinon}, and D.~G. {Caldwell},
  ``Probabilistic learning of torque controllers from kinematic and force
  constraints,'' in \emph{IEEE/RSJ International Conference on Intelligent
  Robots and Systems (IROS)}, Oct 2018, pp. 1--8.

\bibitem{Koutras2019}
L.~Koutras and Z.~Doulgeri, ``{A correct formulation for the Orientation
  Dynamic Movement Primitives for robot control in the Cartesian space},'' in
  \emph{3rd Conference on Robot Learning (CoRL)}, Osaka, 11 2019.

\bibitem{Zeestraten2017}
M.-J. {Zeestraten}, I.~{Havoutis}, J.~{Silv\'erio}, S.~{Calinon}, and D.-G.
  {Caldwell}, ``An approach for imitation learning on riemannian manifolds,''
  \emph{IEEE Robotics and Automation Letters}, vol.~2, pp. 1240--1247, 2017.

\bibitem{Schneider2010a}
M.~{Schneider} and W.~{Ertel}, ``Robot learning by demonstration with local
  {G}aussian {P}rocess regression,'' in \emph{2010 IEEE/RSJ International
  Conference on Intelligent Robots and Systems}, Oct 2010.

\bibitem{Umlauft2017}
J.~{Umlauft}, Y.~{Fanger}, and S.~{Hirche}, ``Bayesian uncertainty modeling for
  programming by demonstration,'' in \emph{IEEE International Conference on
  Robotics and Automation (ICRA)}, 2017, pp. 6428--6434.

\bibitem{Jaquier2019}
N.~Jaquier, D.~Ginsbourger, and S.~Calinon, ``Learning from demonstration with
  model-based {G}aussian {P}rocess,'' in \emph{3rd Conference on Robot Learning
  (CoRL), Osaka, Japan}, Oct 2019.

\bibitem{Schulz2018}
E.~Schulz, M.~Speekenbrink, and A.~Krause, ``A tutorial on {G}aussian {P}rocess
  regression: Modelling, exploring, and exploiting functions,'' \emph{Journal
  of Mathematical Psychology}, vol.~85, pp. 1--16, 2018.

\bibitem{Rana2017}
M.-A. Rana, M.~Mukadam, S.-R. Ahmadzadeh, S.~Chernova, and B.~Boots, ``Towards
  robust skill generalization: unifying learning from demonstration and motion
  planning,'' in \emph{1st Conference on Robot Learning (CoRL)}, CA, USA, 10
  2017.

\bibitem{Osa2018}
T.~{Osa}, N.~{Sugita}, and M.~{Mitsuishi}, ``Online trajectory planning and
  force control for automation of surgical tasks,'' \emph{IEEE Transactions on
  Automation Science and Engineering}, vol.~15, pp. 675--691, 2018.

\bibitem{Lang2018}
M.~Lang, M.~Kleinsteuber, and S.~Hirche, ``Gaussian {P}rocess for {6-DoF} rigid
  motions,'' \emph{Autonomous Robots}, vol.~42, no.~6, 2018.

\bibitem{Lang2017}
M.~{Lang} and S.~{Hirche}, ``Computationally efficient rigid-body {G}aussian
  {P}rocess for motion dynamics,'' \emph{IEEE Robotics and Automation Letters},
  vol.~2, no.~3, pp. 1601--1608, July 2017.

\bibitem{Rasmussen2006}
C.-E. Rasmussen and C.-K.-I. Williams, \emph{{Gaussian Processes for Machine
  Learning}}.\hskip 1em plus 0.5em minus 0.4em\relax {The MIT Press}, 2006.

\bibitem{Alvarez2012}
M.-A. Alv\'arez, L.~Rosasco, and N.-D. Lawrence, ``{Kernels for vector-valued
  functions: A review},'' \emph{Foundations and Trends in Machine Learning},
  vol.~4, no.~3, pp. 195--266, Mar 2012.

\bibitem{Liu2018}
H.~Liu, J.~Cai, and Y.-S. Ong, ``{Remarks on Multi-Output Gaussian Process
  Regression},'' \emph{Knowledge-Based Systems}, vol. 144, pp. 102--121, March
  2018.

\bibitem{Goldberg1998}
P.~Goldberg, C.~Williams, and C.~Bishop, ``Regression with input-dependent
  noise: A {G}aussian {P}rocess treatment,'' \emph{Advances in Neural
  Information Processing Systems}, vol.~10, pp. 493--499, Jan 1998.

\bibitem{Kersting2007}
K.~Kersting, C.~Plagemann, P.~Pfaff, and W.~Burgard, ``Most-likely
  {H}eteroscedastic {G}aussian {P}rocess regression,'' in \emph{ACM
  International Conference Proceeding Series}, vol. 227, Jan 2007, pp.
  393--400.

\bibitem{Senin2008}
P.~Senin, ``{Dynamic Time Warping algorithm review},'' Information and Computer
  Science Department, Univerity of Hawaii at Manoa, Honolulu, USA, Tech. Rep.,
  Dec 2008.

\bibitem{Deisenroth2015a}
M.~Deisenroth and J.-W. Ng, ``Distributed {G}aussian {P}rocesses,'' in
  \emph{32nd International Conference on Machine Learning (ICML)}, vol.~37,
  July 2015, pp. 1481--1490.

\bibitem{Bilj2018}
H.-L. Bilj, ``{LQG and Gaussian Process Techniques for Fixed-Structure Wind
  Turbine Control},'' PhD Dissertation, Delft University of Technology, The
  Netherlands, Tech. Rep., Oct 2018.

\bibitem{Kronander2016}
K.~{Kronander} and A.~{Billard}, ``Stability considerations for variable
  impedance control,'' \emph{IEEE Transactions on Robotics}, vol.~32, no.~5,
  pp. 1298--1305, Oct 2016.

\bibitem{Kim2004}
D.~Kim, J.-H. Kang, C.-S.~H. CS., and G.-T. Park, ``Mobile robot for door
  opening in a house,'' in \emph{Knowledge-Based Intelligent Information and
  Engineering Systems (KES), Part II}, Sept 2004, pp. 596--602.

\end{thebibliography}

\end{document}